\documentclass[research]{fcs}
\usepackage{colortbl}  
\usepackage{adjustbox}
\usepackage{multirow}
\usepackage{algorithm}
\usepackage{algorithmic}
\usepackage{makecell}
\usepackage{booktabs}
\usepackage{pifont}
\usepackage{microtype}
\usepackage{placeins}
\usepackage{enumitem}
\usepackage{hhline} 

\newtheorem{assumption}{Assumption}

\title{MiMu: Mitigating Multiple Shortcut Learning Behavior of Transformers}
\author[1]{Lili Zhao}
\author*[1]{Qi Liu}
\author[1]{Wei Chen}
\author[1]{Liyi Chen}
\author[1]{Ruijun Sun}
\author[2]{Min Hou}
\author[1,3]{Yang Wang}
\author[3]{Shijin Wang}
\address[1]{State Key Laboratory of Cognitive Intelligence, University of Science and Technology of China, Hefei 230026, China}
\address[2]{Hefei University of Technology, Hefei 230009, China}
\address[3]{iFLYTEK AI Research (Central China), iFLYTEK Co., Ltd, Hefei 230088, China}
\fcssetup{
  received       = {month dd, yyyy},
  accepted       = {month dd, yyyy},
  corr-email     = {qiliuql@ustc.edu.cn},
}
\begin{abstract}
  Empirical Risk Minimization (ERM) models often rely on spurious correlations between features and labels during the learning process, leading to shortcut learning behavior that undermines robustness generalization performance. 
  Current research mainly targets identifying or mitigating a single shortcut; however, in real-world scenarios, cues within the data are diverse and unknown. 
  In empirical studies, we reveal that the models rely to varying extents on different shortcuts. Compared to weak shortcuts, models depend more heavily on strong shortcuts, resulting in their poor generalization ability. 
  To address these challenges, we propose \textbf{MiMu}, a novel method integrated with Transformer-based ERMs designed to \textbf{Mi}tigate \textbf{Mu}ltiple shortcut learning behavior, which incorporates self-calibration strategy and self-improvement strategy.
  In the source model, we preliminarily propose the self-calibration strategy to prevent the model from relying on shortcuts and make overconfident predictions. 
  Then, we further design self-improvement strategy in target model to reduce the reliance on multiple shortcuts. 
  The random mask strategy involves randomly masking partial attention positions to diversify the focus of target model other than concentrating on a fixed region. Meanwhile, the adaptive attention alignment module facilitates the alignment of attention weights to the calibrated source model, without the need for post-hoc attention maps or supervision. 
  Finally, extensive experiments conducted on Natural Language Processing (NLP) and Computer Vision (CV) demonstrate the effectiveness of MiMu in improving robustness generalization abilities.
\end{abstract}

\begin{document}

\begin{figure}[t]
  \centering
  \includegraphics[width=0.48\textwidth]{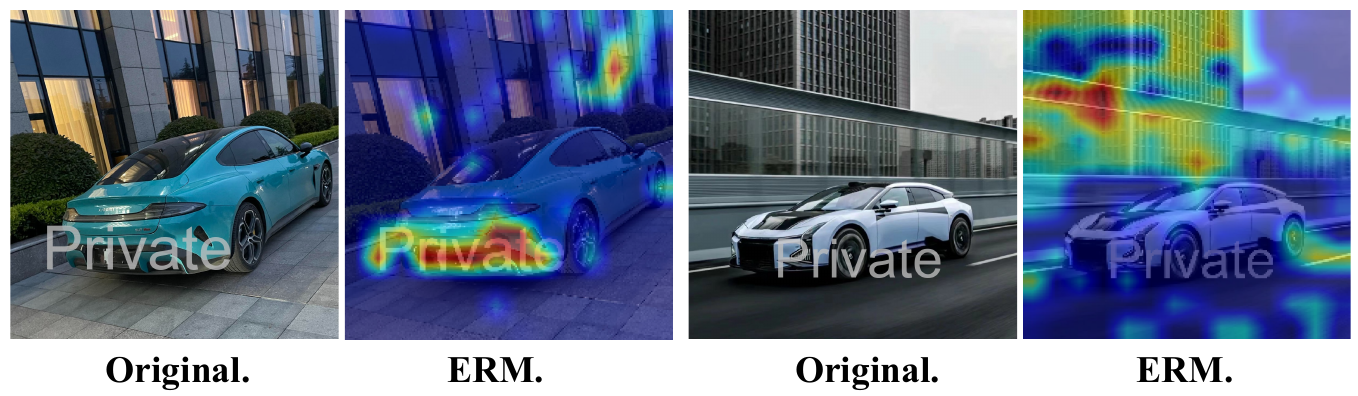}
  \caption{Saliency maps of ViT-Large show that it is unknown and diverse what kind of shortcuts (watermark or background) the model relies on in real-world scenarios.}
  \label{fig:case_erm}
\end{figure}

\section{Introduction}
\label{sec:intro}

In deep learning systems, Empirical Risk Minimization (ERM) \cite{erm} optimizes training parameters by minimizing the loss function on training data, thereby achieving effective learning results \cite{transformer,vit}.
Nevertheless, the paradigm could potentially lead the model to learn easy-to-master patterns during the training process \cite{NMI20,nips2023don}. 
Such patterns often reflect spurious correlations between certain features and labels, known as shortcut learning behavior \cite{NMI20,hupkes2023taxonomy}, resulting in the model performing well on Independent and Identically Distributed (IID) data, but struggling with poor generalization on Out-Of-Distribution (OOD) data\cite{zhao2024comi,zhao2025evaluating}.

Existing works mainly depend on identifying or inferring recognized shortcuts to mitigate shortcut learning \cite{acl23,du2023shortcut}. For instance, in Natural Language Processing (NLP), investigations have uncovered specific shortcuts, such as spurious correlations between negation words and the label ``contradiction'' \cite{naik2018stress}, or word overlap with label ``entailment'' \cite{hans}. 
To address that, researchers \cite{poe,cr,samplerew} adjust the loss function by giving specific known shortcut to correct the conditional distribution of the model's predictions.
Similarly, in Computer Vision (CV), scholars have identified shortcuts like correlations between objects and backgrounds \cite{bg-iclr21,asgari2022masktune} or textures \cite{texture-iclr18}. 
These shortcuts are mitigated by re-weighting \cite{reweight-iclr19,cv-mixup} or targeted augmentation \cite{bg-iclr21}, which are applied based on prior knowledge of shortcuts.
Although some methods \cite{pseudo-LfF,pseudo-EIIL} attempt to infer pseudo shortcut labels in the absence of prior knowledge, they are founded on strong assumptions about shortcut learning process, only adapting to an individual shortcut.

However, in real-world scenarios, due to the complexity and diversity of data, the specific shortcuts a model relies on are often diverse. As shown in Fig. \ref{fig:case_erm}, the model may rely on watermark or background shortcut in the image classification task, highlighting the need for a more robust method that can adapt to the complex real-world conditions.

To this end, we propose a novel method, \textbf{MiMu}, designed to \textbf{Mi}tigate \textbf{M}ultiple shortcut learning behavior in learning systems. 
Particularly, in our empirical investigation, we find that the model has different levels of reliance on different shortcuts. Moreover, compared to weak shortcuts, the model relies more on strong shortcuts, leading to its poorer generalization ability.
Meanwhile, when the model captures multiple shortcuts, its performance typically lies between the performances observed when relying on only a single shortcut. It implies that without prior knowledge of which shortcuts are strong or weak, eliminating a certain shortcut may cause the model to rely more heavily on strong shortcuts.

To address these issues, we design self-calibration and self-improvement strategies that are achieved by source model and target model respectively integrated with Transformer-based ERMs, to jointly mitigate multiple shortcut learning behavior.
Specifically, since models often rely on shortcut features accompanying  overconfident predictions \cite{hermann2020shapes,du2023shortcut,naacl21}, we first design self-calibration strategy in the source model to better align the model's prediction with its confidence, thereby ensuring prediction accuracy and reducing over-reliance on strong shortcuts.
Then, self-improvement strategy is proposed in the target model to force the model to consider more features and further reduce model's reliance on multiple shortcuts.
Since the shortcuts captured by the model remain unknown during training, we leverage the internal attention weights as the model prediction's explanations on tokens in text or patches in images, and introduce random mask strategy to prevent the target model from focusing on a fixed region that facilitates shortcut learning. This involves randomly masking partial positions in the attention weights, thereby forcing the target model to diversify its focus.

However, models may capture strong shortcuts in the remaining locations outside the mask. The source model can serve as a reference model to provide stable attention weights during training and soft outputs, which supports the adaptive attention alignment module in self-improvement strategy, aligning the attention weights to the source model (calibrated source model promotes richer attention weights) without extra supervision. 
Besides, we further develop the post-hoc calibration approach to synchronize the target model's output probabilities with its confidence levels, preventing the learning process of the target model from shifting.

Due to the singularity of shortcuts in existing CV datasets, we add different forms of watermarks commonly found in real-world data, expand the training set and construct variants of OOD test sets.
Extensive experiments across three tasks, including \textbf{Natural Language Inference} (NLI) and \textbf{Understanding} (fact verification) in NLP, and \textbf{Image Classification} (IC) in CV suggest that MiMu can significantly improve the performance on OOD samples while maintaining IID accuracy based on Transformer-based ERMs.

\begin{figure*}[t]
    \centering
    \includegraphics[width=0.9\textwidth]{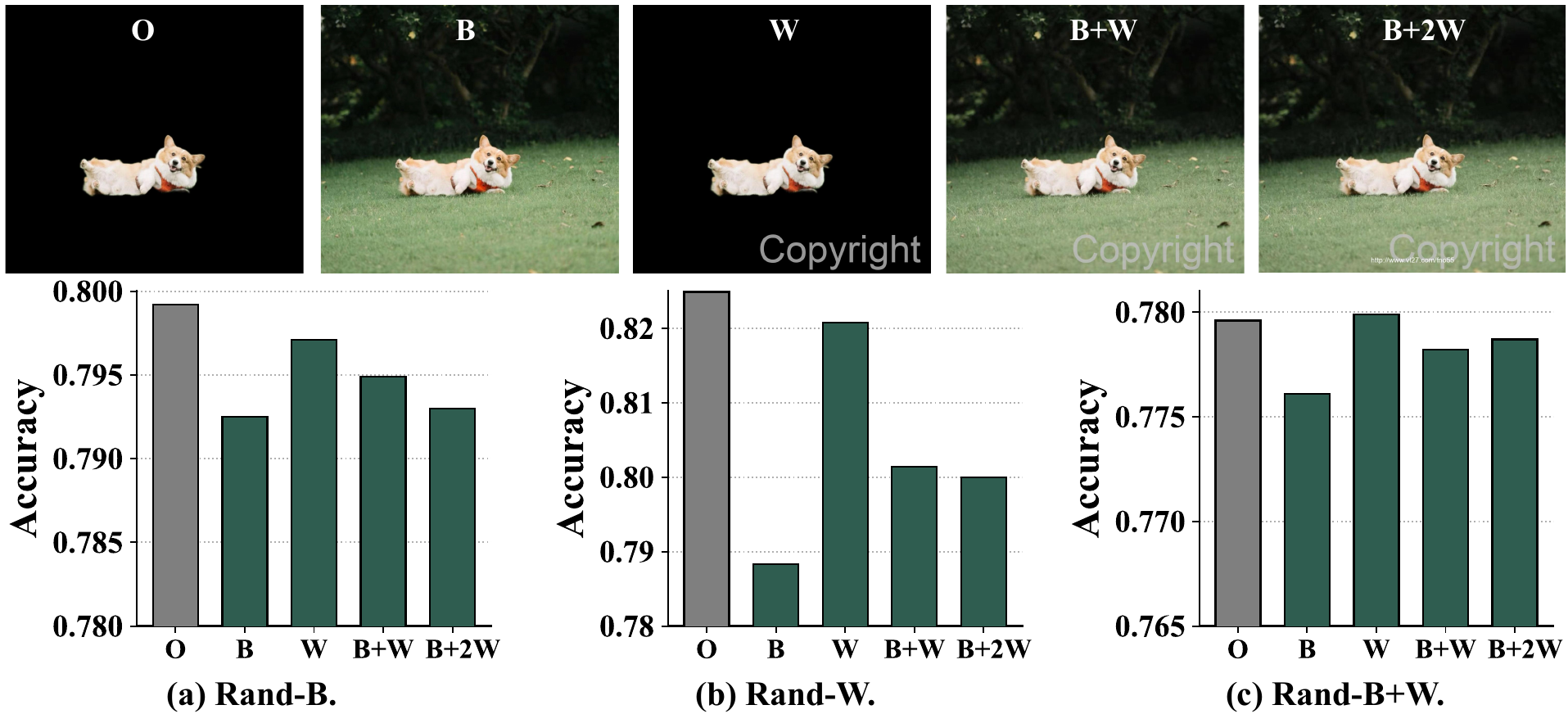}
    \caption{We employ 5 variants of ImageNet-9 training data (\textbf{O}bject, \textbf{B}ackground, \textbf{W}atermark, \textbf{B}ackground\textbf{+}\textbf{W}atermark, \textbf{B}ackground\textbf{+}\textbf{2W}atermark), including multiple combinations of background and watermark as shown in the top row. We evaluate the performance on the three variants (\textbf{Rand-B}, \textbf{Rand-W}, \textbf{Rand-B+W}) of OOD test sets (Mixed-Rand).}
    \label{fig:ei}
 \end{figure*}
 
\section{Related Work}
\label{sec:rw}
\subsection{Determining Shortcut Learning}
Shortcut learning refers to the phenomenon that decision rules rely heavily on non-robust features as shortcuts, performing well on standard benchmarks but failing to transfer to more challenging testing conditions, such as real-world scenarios \cite{NMI20, hupkes2023taxonomy}. 
Several studies \cite{jin2020bert-attack,naik2018stress} have identified key instances of shortcut learning behavior in \textbf{text data}. 
Lexical shortcut \cite{naacl21,acl23} exhibited a high co-occurrence correlation with labels such as stop words, numbers, and negation words. Overlap shortcut \cite{hans,laioverlap-acl21} utilized the lexical overlap, subsequence or constituent biases between pairs of paragraphs other than the underlying information. Besides, other shortcuts, such as position \cite{yuan2024llms,llm-acl23} and style \cite{nlp-style} were also encountered in NLI tasks.
In \textbf{image data}, \cite{texture-iclr18} revealed that standard models classified images based on the object texture other than the real shape. Other research \cite{bg-eccv18,bg-iclr21,color-cvpr23} found object recognition models tended to depend on signals from image backgrounds or object colors. Recently, \cite{liwhac-cvpr23} investigated the phenomenon of multiple shortcuts, which demonstrated shortcuts come in multiples where mitigating one amplifies others. 
In \textbf{multi-modal data}, \cite{ge2023learning,liang2021lpf} disclosed that models learned shortcut that tended to over-encode some informative modalities while neglecting other modalities.

\subsection{Addressing Shortcut Learning} 
It is essential to enhance the OOD generalization and robustness of models \cite{hupkes2023taxonomy}, while preserving  predictive performance on IID samples \cite{du2023shortcut,lastlayer-iclr23}. 
\textbf{In NLP}, various methods employed traditional data augmentation \cite{dataaug-tacl20,dataaug-si2021} and model-centric approaches to counteract shortcuts.
In model-centric methods, they trained bias-only model and debiased model to adjust the loss function based on known and pre-existing bias. 
Specifically, Sample Re-weighting (SR) \cite{samplerew,emnlp20} assigned higher training weights on hard training samples. Based on the known shortcut, Product Of Experts (POE) \cite{poe-unlearn,poe,ensemble} trained a debiased model by combining it with a bias-only model. Confidence Regularization (CR) \cite{cr, naacl21} promoted higher uncertainty (lower confidence) in the debiased model's predictions for biased samples. 
The main implementations and loss functions of these methods are listed in Section \ref{exp-methods}.
Recently, \cite{acl23} introduced human attention to shift attention from shortcut features to genuine features, 
\cite{contrastive-nips21} incorporated contrastive learning and made augmentation with counterfactual samples \cite{counterfactuals-aaai21} to guide learning.

\textbf{In CV}, without the prior knowledge of shortcuts, standard data augmentation, and regularization techniques \cite{cv-mixup,cv-cutout,cv-cutmix,cv-augmix,hong2025regularization} could improve OOD robustness. 
With the given shortcuts, some researchers designed targeted augmentation to make mitigation, such as texture \cite{texture-iclr18} and background augmentation \cite{bg-iclr21}. Some methods also employed shortcut labels for mitigation, which involved techniques like re-weighting \cite{reweight-iclr19} and resampling training data \cite{lastlayer-iclr23}. In the absence of known shortcut, some studies \cite{pseudo-LfF,pseudo-JTT,pseudo-EIIL,pseudo-DebiAN} attempted to infer pseudo shortcut labels, which mainly established on a single shortcut. \cite{zhao2024comi} proposed COMI with information separation to encourage models rely less on shortcuts and focus on real features both in NLP and CV domains.
Moreover, invariant learning was proposed to filter out the variant spurious correlations and learn invariant relationships, which always based on strong assumptions \cite{arjovsky2019invariant,koyama2020out,chang2020invariant} and heavily rely on multiple environments that have to be explicitly provided or pre-defined in the training dataset \cite{nips22zin}.

Although most works ignored to specify shift source and shift type they focused on \cite{hupkes2023taxonomy}, in this paper, \textbf{to unify the settings of NLP and CV, we focus more on generated shifts}, i.e., training models on natural data and making post-hoc evaluations on generated OOD sets.
The shift type we are concerned with is \textbf{conditional feature distribution shift}, where some conditional features based on label $y$ change in OOD setting but do not affect $y$. 
Most papers in CV \cite{liwhac-cvpr23,asgari2022masktune} also involve class imbalance problem, i.e., long-tailed distribution where a small portion of the distribution in the training set remains consistent with the OOD testing set, which does not occur in NLP datasets.
We provide a detailed discussion and comparisons about shift definition, shift type, shift source in shortcut learning with other shift types in Appendix \ref{append:sl-debias}.

\section{Empirical Investigation}
\subsection{Findings on Multiple Shortcuts}
\label{investigation}
In real-world scenarios, specific features in text or images are strongly correlated with labels and remain unknown. 
To evaluate whether the model captures different shortcuts during the training process, we employ five settings on training set (ImageNet-9) as illustrated in top row of Fig. \ref{fig:ei}\footnote{We conduct training under 5 different settings: \textbf{O} (Objects only), \textbf{B} (Backgrounds included), \textbf{W} (Watermarks included), \textbf{B+W} (Backgrounds and Watermarks), \textbf{B+2W} (Backgrounds and Double Watermarks-``copyright'' and urls).}. 
To evaluate how strongly the model relies on the \textbf{Background} or \textbf{Watermark} in OOD setting, we adopt 3 variants of OOD set (Mixed-Rand) to evaluate how strongly the model relies on the \textbf{Background} or \textbf{Watermark}, which contains Mixed-Rand-: \textbf{B} (random background of a random class), \textbf{W} (random watermark of remaining classes), \textbf{B+W} (overlay of random background and watermark).

To explore different shortcut learning behaviors of the model, \cite{liwhac-cvpr23} focused on the fully generated dataset-UrbanCars with subpopulation shift \cite{yang2023sub}, and the position, size of generated shortcuts (object and co-occurrence) are fixed.
To fit the diversity of real scenes and the unknown shortcuts during training, we diversify the font, size and position of added watermarks to model the real world scenarios. Besides, we train on natural data (ImageNet-9) and evaluate on generated data (Mixed-Rand) to focus on spurious correlations of possible features and labels during training.

As shown in the experimental results in the bottom row of Fig. \ref{fig:ei}, we find that the model trained solely on objects exhibits strong performance overall on OOD datasets, enabling the model to focus on underlying features of objects and enhancing its generalization abilities.
In the setting with shortcuts, the model trained on the background exhibits the poorest performance, suggesting a strong reliance on backgrounds (\textbf{strong shortcuts}). 
Conversely, the model trained on the watermark demonstrates preferable performance, indicating a weak reliance on watermarks (\textbf{weak shortcuts}).
Besides, when we introduce multiple shortcuts, \textbf{the performance of both B+W and B+2W is somewhere between B and W}, the model reduces its reliance on the background, resulting in its performance falling between the effects of individual shortcuts.
Therefore, we can draw two assumptions:

\begin{figure*}[t]
    \centering
    \includegraphics[width=\textwidth]{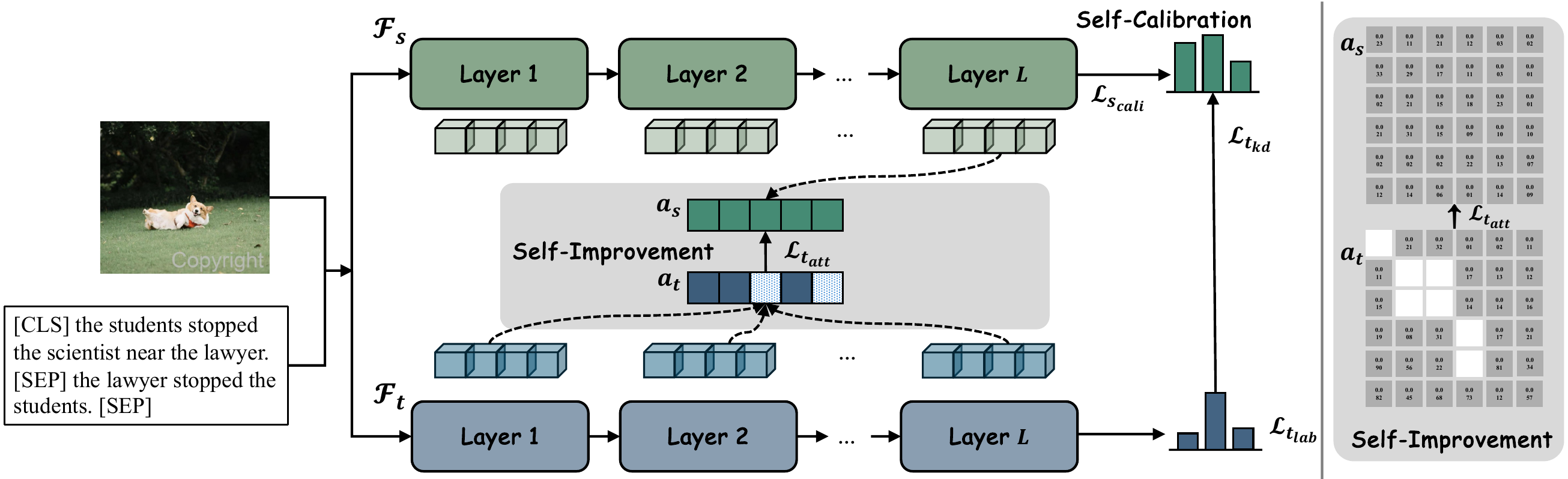}
    \caption{The self-calibration strategy in source model ($\mathcal{F}_s$) and self-improvement strategy in target model ($\mathcal{F}_t$) of MiMu jointly help mitigate multiple shortcut learning behavior in NLP and CV. Layer denotes the Transformer layer/block in ERM model.}
    \label{att_align}
\end{figure*} 

\begin{assumption}
    \label{ass:1}
    The model has different levels of reliance on different shortcuts, which we call strong shortcuts and weak shortcuts. Compared to weak shortcuts, the model exhibits more reliance on strong shortcuts, leading to poorer generalization ability.
\end{assumption}

\begin{assumption}
    \label{ass:2}
    The performance of the model under multiple shortcuts often falls between the performance under a single shortcut.
\end{assumption}

\subsection{Problem Definition}
\label{sec:pd}

In this paper, given input $x$ towards a specific target variable $y$, the mapping function $f$ encompasses all important features to make predictions, i.e., $y = f(x)$.
However, within $f$ there might exist a set of shortcuts $s = \{s_1, s_2, ..., s_k\}$. 
The presence of $s$ within $f$ can be either existent or non-existent. Therefore, our goal is to find a new mapping function $g$ that is as close as possible to $f$, but without the influence of $s_i$. It can be represented as an optimization problem:

\begin{equation}
\min_{g}(\|g - f\|^2 + \lambda \sum_{i=1}^k I(s_i \in g)).
\end{equation}

Here, $\|g - f\|^2$ represents the difference between $g$ and $f$, $I(s_i \in g)$ is an indicator function that is 1 if $g$ includes $s_i$ and 0 otherwise, and $\lambda$ is a hyperparameter that balances the two objectives.
In this optimization problem, we hope to find a $g$ that is as close as possible to $f$ (i.e., minimize $\|g - f\|^2$) and does not include any $s_i$ (i.e., minimize $\sum_{i=1}^k I(s_i \in g)$). 

To this end, we investigate the problem with ERM models through Natural Language Inference (NLI) and Paraphrase Identification (PI) in NLP, as well as Image Classification (IC) in CV. 
To unify the shift settings, we concentrate on the \textbf{conditional feature distribution shift}, i.e. $P_{tr}(X | Y) \neq P_{te}(X | Y)$ and $P_{tr}(Y) = P_{te}(Y)$, where $tr$ and $te$ represent the training and OOD test sets respectively.
Specifically, given two sentences (i.e. $x_i$ and $x_j$), our goal is to learn a classifier $\xi$ that can predict the label $y$ (e.g., entailment, contradiction, or neutrality in NLI; duplicate or non-duplicate in PI). 
In IC, given an image $x$, the model predicts the label $y$ (e.g., dog).

\begin{figure*}[t]
    \centering
    \includegraphics[width=0.7\textwidth]{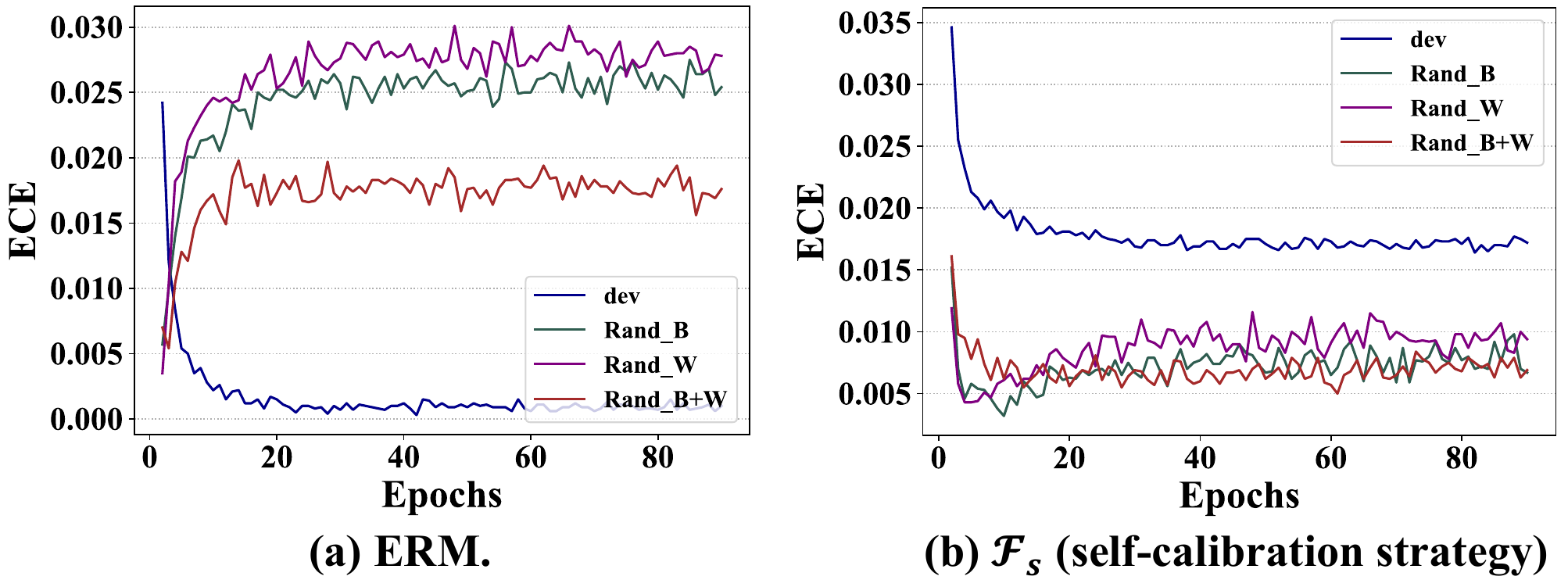}
    \caption{The ECE scores \cite{calibration} of ERM model and source model ($\mathcal{F}_s$) on the three OOD sets. A lower ECE score indicates better calibration. ERM model often relies on shortcuts, accompanying confident predictions (higher ECE scores in OOD sets).}
    \label{ece-i9} 
\end{figure*} 

\section{MiMu: Mitigate Multiple Shortcut Learning Behavior}
To mitigate multiple shortcut learning behavior of ERMs, we mainly propose self-calibration and self-improvement strategies in MiMu as shown in Fig. \ref{att_align}. The former is completed by source model \textbf{$\mathcal{F}_s$}, while the latter is completed by target model \textbf{$\mathcal{F}_t$}. \textbf{$\mathcal{F}_s$ and $\mathcal{F}_t$ share the same structure following self-distillation approach}\footnote{Distilling at fine-grained levels, such as attention weights and features, can lead to misalignment of model capabilities; however, self-distillation avoids this phenomenon.}.

\subsection{Self-Calibration Strategy}

ERM model often captures strong shortcuts rapidly during the initial epochs of training and providing overconfident predictions \cite{hermann2020shapes,du2023shortcut,cr,naacl21,bg-iclr21} as shown in Fig. \ref{ece-i9} (a).
To tackle this, we need self-calibration\footnote{Calibration defines that a credible and reliable model should ensure that the model's probability estimates (confidence) match with the actual probability of correctness.} to align the model's outputs with its confidence in $\mathcal{F}_s$ to prevent the model from becoming overconfident and relying on shortcuts preliminarily. Besides, $\mathcal{F}_s$ needs to provide reliable logistic information and attention weights for $\mathcal{F}_t$ to maintain accuracy.

However, existing confidence calibration methods fail to meet these requirements. Specifically, 

\begin{itemize}
    \item  Post-hoc calibration (platt scaling) \cite{calibration} adjusts the prediction probability (e.g. softmax output) after training, without helping attention distribution during training. 
    \item Confidence regularization (label smoothing) \cite{labelsmoothing} reduces model confidence by regularization in training, but impairs distillation. Target models perform poorly due to information loss in logits when source model is trained with label smoothing \cite{muller2019does}.
\end{itemize}

Therefore, we design the simple but effective calibration loss to \textbf{align the model's outputs with its confidence and ensure prediction accuracy.}
The mean squared difference with weight $\lambda_c$ and cross-entropy between the predicted outputs $\hat{y_s}$ of $\mathcal{F}_s$ and the true distribution $y$ are adopted as the calibration loss, effectively penalizing samples with high confidence but incorrect predictions as shown in Fig. \ref{ece-i9} (b), where $K$ is the number of classes:

\begin{equation}
    \label{eq1}
     \mathcal{L}_{s_{\text {cali}}} =  \lambda_c \sum_{k=1}^{K} (\hat{y_{s_k}} - y_{k})^2  -\sum_{k=1}^{K} y_{k} \log(\hat{y}_{s_k}).
\end{equation}

Besides, to understand the decision-making process during model prediction and help further reduce the model's reliance on multiple shortcuts within the self-improvement strategy, we derive the intrinsic attention weights from the \textbf{last layer of the transformer-based $\mathcal{F}_s$.}
We then compute the average of these attention weights across all sequence positions, denoted as $a_s=\left\{a_{s_i}\right\}_{i=1}^{l}$. Here, $a_{s_i}$ signifies the attention weight assigned to the $i$-th position in $\mathcal{F}_s$, $l$ denotes the sequence length\footnote{Sequence length refers to the number of tokens in NLP, or the number of patches plus one class token in CV (e.g., a 14x14 grid of patches plus one class token for a 224x224 input image). For each attention head, the matrix size is 197×197.}.

\subsection{Self-Improvement Strategy}

To further reduce the model's reliance on multiple shortcuts, we develop a self-improvement strategy for $\mathcal{F}_t$. This strategy begins with the implementation of a distillation loss, designed to ensure that the softened output distribution $P_t$ of $\mathcal{F}_t$ closely mirrors the output distribution $P_s$ of $\mathcal{F}_s$. In addition, we employ $\mathcal{L}_{t_{\text{lab}}}$ to align outputs $\hat{y}_t$ with the true distribution $y$:

\begin{equation} 
    \label{eq2}
    \mathcal{L}_{t_{\text {kd}}} = T^2 \cdot D_{\text {KL}}(P_t || P_s), \\
    \mathcal{L}_{t_{\text {lab}}} = -\sum_{k=1}^{K} y_{k} \log(\hat{y}_{t_k}),
\end{equation}
where temperature parameter $T$ encourages $\mathcal{F}_t$ to consider a broader range of possibilities, which can help $\mathcal{F}_t$ learn more generalized representations, especially when the $\mathcal{F}_s$ may learn shortcuts to make overconfident predictions. $D_{\text {KL}}$ represents Kullback-Leibler (KL) divergence of $P_t$ and $P_s$.

In Assumptions \ref{ass:1} and \ref{ass:2} of Section \ref{investigation}, we observe that the model has different levels of reliance on different shortcuts; multiple shortcuts can lead to the model's performance falling between the performance of individual shortcuts. 
\textbf{During the process of eliminating shortcuts, it is possible to eliminate weak shortcuts while retaining strong ones.}
To tackle this, we first introduce a random masking strategy to selectively obscure $N$ positions, which forces $\mathcal{F}_t$ to focus on broader locations instead of focusing on a fixed region to learn shortcuts.
Then, without post-hoc attention map generation \cite{asgari2022masktune,zhao2024comi} or additional supervision \cite{li2018tell,acl23}, we achieve adaptive attention alignment module to align the attention weights with those of $\mathcal{F}_s$ (calibrated source model promotes richer attention weights), preventing the remaining locations outside the mask from containing strong shortcuts.
We compute \textbf{the average attention weights $a_t = \left\{a_{t_i}\right\}_{i=1}^{l}$ across all layers} in $\mathcal{F}_t$, align the attention weights $a'_t$ of the remaining $l-N$ positions with \textbf{those attention weights of last layer} $a'_s$ of $\mathcal{F}_s$, promoting a more robust and consistent learning process. 

This entire process is adaptively learned by $\mathcal{F}_t$. To quantify this alignment, we propose $\mathcal{L}_{t_{\text{att}}}$ to compute the top $M$ difference values, where $M$ represents the number of top differences selected, and $\text{Norm}$, denoting min-max normalization, helps to make the feature distribution comparable:

\begin{equation} 
    \label{eq3}
\mathcal{L}_{t_{\text {att}}} =  \frac{1}{M} \sum_{i=1}^{M} | \text{Norm}(a'_{t_{i}}) - \text{Norm}(a'_{s_{i}}) |.
\end{equation}

\begin{algorithm}[t]
    \caption{Proposed MiMu.}
    \label{algorithm}
    \begin{algorithmic}[1]
    \REQUIRE Data in $D$, ERM. 
    \WHILE{training stage}
        \STATE Training Source model $\mathcal{F}_s$ by self-calibration strategy using Eq. \ref{eq1}, obtain $a_s$, $P_s$, Fix parameters;
        \STATE Training Target model $\mathcal{F}_t$ using Eq. \ref{eq2}, obtain $a_t$;
        \STATE Self-improvement strategy using Eq. \ref{eq3};
        \STATE Training logistic regression model for $\mathcal{F}_t$ calibration using Eq. \ref{eq4}
        \STATE Post-hoc Calibration for $\mathcal{F}_t$ using Eq. \ref{eq5}
    \ENDWHILE
    \WHILE{inference stage}
        \STATE  Make the prediction based on values of the first $K$ categories through $\mathcal{F}_t$:\\
        $\hat{y}_i = \arg\max_{j \in [1, \cdots, K]} p_{t_{i,j}}$
    \ENDWHILE    
    \end{algorithmic}
\end{algorithm}

Additionally, to prevent the model learning process from shifting, where the $\mathcal{F}_t$ aligns with the shortcuts that $\mathcal{F}_s$ may learn, we adopt a post-hoc calibration approach to synchronize the $\mathcal{F}_t$'s output probabilities with its confidence levels. 
We calibrate $\mathcal{F}_t$ through the training of a logistic regression model for each category. The goal is to optimize the parameters $A_k$ and $B_k$ of the logistic regression model for category $k$, thereby ensuring more accurate and reliable probability outputs:

\begin{equation}
    \label{eq4}
\min_{A_k, B_k} (- \left[ y_{k} \log(p_{k}) + (1 - y_{k}) \log(1 - p_{k}) \right]), 
\end{equation}
where $p_{k} = \frac{1}{1 + e^{A_k f_{k} + B_k}}$ is the probability predicted by the logistic regression model, $f_{k}$ is the score output of $\mathcal{F}_t$ in $k$-th category, $y_{k}$ indicates whether the sample belongs to category $k$ (0 or 1). 
Then, we utilize the optimized parameters $A_k$ and $B_k$ to convert $f_{k}$ and make normalization to get the final output distribution $P_t$ of target model $\mathcal{F}_t$:

\begin{equation}
    \label{eq5}
    \begin{split}
P_t(y = k | f_{k}) = \frac{1}{1 + e^{A_k f_{k} + B_k}}, \\
P_t(y = k | f_{ik}) = \frac{P_t(y = k | f_{ik})}{\sum_{j=1}^{K} P_t(y = j | f_{ij})}.
    \end{split}
\end{equation}

\subsection{Training and Optimization}
Through the process of the self-calibration and self-improvement strategies, we aim to address multiple shortcut learning issues without targeted augmentation. By optimizing the defined losses mentioned earlier, the target model $\mathcal{F}_t$ can effectively learn more generalized representations. 
The optimization objectives could be precisely defined as follows:
\begin{equation}
    \mathcal{L}=\mathcal{L}_{t_{\text {lab}}}+\lambda_1\mathcal{L}_{t_{\text {kd}}}+\lambda_2\mathcal{L}_{t_{\text {att}}},
\end{equation}
where $\lambda_1$ and $\lambda_2$ are regularization hyper-parameters.
The algorithm of MiMu is shown in Algorithm \ref{algorithm}.

\section{Experiments}
\label{expri}
To assess the effectiveness of MiMu, we conduct a comprehensive analysis involving three distinct tasks and four datasets spanning NLP and CV fields. We compare MiMu with several benchmark models and assess the contributions of our strategies. Moreover, experiments on calibration, saliency, representation visualization, hyperparameter $N$ are performed to assess its performance thoroughly.

\subsection{Experimental Setup}
\label{ex-setup}

\begin{table*}
  \centering
  \caption{The statistics of the datasets in NLP and CV.}
  \resizebox{\textwidth}{!}{
  \begin{tabular}{c|cccc|ccc}
      \hline
      \hline
      \multirow{2}*{\textbf{NLP}} &  \cellcolor[gray]{0.92}\textbf{MNLI}-Train \cite{MNLI} & \cellcolor[gray]{0.92}Match & \cellcolor[gray]{0.92}Mismatch   & \cellcolor[gray]{0.92}Hans & \cellcolor[gray]{0.92}\textbf{QQP}-Train \cite{qqp} & \cellcolor[gray]{0.92}Dev & \cellcolor[gray]{0.92}PAWS \\
      ~ &  392,702 & 9,815 & 9,832 & 30,000 & 363,846   & 40,430  & 8,000   \\
      \hline
      \hline
      \multirow{2}*{\textbf{CV}}  & \cellcolor[gray]{0.92}\textbf{IN-9-W}-Train \cite{bg-iclr21}  & \cellcolor[gray]{0.92}Dev & \cellcolor[gray]{0.92}Mixed-Rand-(B/W/B+W) & \cellcolor[gray]{0.92}- & \cellcolor[gray]{0.92}\textbf{IN-200-W}-Train \cite{imagenet} & \cellcolor[gray]{0.92}Dev     & \cellcolor[gray]{0.92}IN-(B/R/R+W)  \\
      ~ & 45,395      & 4,185   & 4,185  & - & 100,000     & 10,000  & 30,000 \\
      \bottomrule
  \end{tabular}}
  \label{dataset}
\end{table*}

\subsubsection{Datasets}
To validate the performance and robustness generalization capabilities of MiMu on IID and OOD samples, we conduct extensive experiments using following four datasets in NLP and CV. Each dataset has its own training, dev (i.e. IID), and OOD subsets. 
\textbf{In NLP}, we employ datasets of MNLI (Multi-Genre Natural Language Inference) \cite{MNLI} in NLI and QQP (Quora Question Pairs)\cite{qqp} in PI, the former involves predicting the relationship (entailment, contradiction or neutrality) between premise and hypothesis, and the latter discerns whether questions pairs are semantically duplicated or non-duplicated\footnote{The sentences separated by the "[SEP]" symbol, are fed into the model as a concatenated pair in both tasks.}.
In \textbf{MNLI}, \underline{MNLI-match} and \underline{MNLI-mismatch} are adopted as IID datasets, while the adversarial dataset \underline{Hans} \cite{hans} is utilized to assess robustness generalization on lexical overlap, subsequence, constituent shortcuts.
In \textbf{QQP}, we employ the \underline{PAWS} \cite{paws} dataset as OOD test set to assess performance of models on lexical overlap shortcuts.

\textbf{In CV}, 
to fully evaluate \textbf{the various shortcuts of the model that may be captured during training} and its performance, in addition to existing shortcuts such as background \cite{bg-iclr21} and texture \cite{texture-iclr18} found in natural datasets, we enhance the existing training sets by adding different watermarks to each class, while also reflecting real-world image conditions.
Unlike the synthetic dataset in \cite{liwhac-cvpr23}, the size and position of shortcuts (objects and co-occurrence relationships) are fixed. The model relies more on fixed relationships of shortcuts, rather than reflecting the shortcut learning behavior of the model in real scenarios. \textbf{However, we add watermarks in different sizes, different colors, and positions on the images to pay more attention to the diversity of real-world scenes.}

Specifically, to evaluate the potential shortcut learning behaviors of the model, we employ different training sets with their validation sets as IID datasets, and adopt multiple variants of OOD datasets to assess various shortcut learning behaviors.
In ImageNet-9 dataset \cite{bg-iclr21}, we put different watermarks about image protection\footnote{Words about image protection such as ``Copyright'', ``Respect'', ``Original'', ``Protect'', 
``Private'', ``License'', ``Ownership'', ``Watermark'', ``Allowed''.}, \textbf{IN-9-W}, as the training set.
For OOD evaluation, we employ three OOD variants on Mixed-Rand: \underline{Rand-B}, \underline{Rand-W} and \underline{Rand-B+W} as illustrated in Section \ref{investigation}, to assess the extent to which the model captures background, watermark, and dual-shortcuts, respectively.
In the ImageNet-200 dataset derived from ILSVRC \cite{imagenet}, we construct 200 web URLs as image watermarks to simulate real-world scenarios, resulting in \textbf{IN-200-W} as training set\footnote{The dataset shares the same image categories as IN-R. Natural images include possible shortcuts: background, texture, and watermarks that we add.}. For OOD evaluation, we utilize three variants of OOD sets to assess the impact of background, texture, and watermarks: firstly by replacing backgrounds in the validated set with random backgrounds from other classes (\underline{IN-B}), then by adopting ImageNet-R \cite{in-r} (\underline{IN-R}) with different textures such as art, cartoon, embroidery etc., and finally by adding random watermarks of other classes on IN-R (\underline{IN-R+W}). 
The detailed statistics of datasets are shown in Table \ref{dataset} and detailed examples are shown in Appendix \ref{append:data}.

\begin{table*}
    \centering
    \caption{The experimental results on the MNLI and QQP in NLP, and their corresponding adversarial datasets (Hans, PAWS), where the results of baselines SR, POE, and CR are reported in \cite{acl23} (dev-m and dev-mm represent the subsets of match and mismatch of MNLI), ``Prior'' stands for these works are built on the prior bias.}
    \resizebox{\textwidth}{!}{
        \begin{tabular}{c|c|ccccc|cccccc}
        \toprule
        \multirow{2}{*}{\textbf{Methods}} & \multirow{2}{*}{\textbf{Prior}} & \multicolumn{5}{c|}{\textbf{MNLI}}  & \multicolumn{6}{c}{\textbf{QQP}} \\
        \cmidrule{3-13}
         ~ & ~ & dev-m & dev-mm & $\Delta^{\uparrow}$ & Hans & $\Delta^{\uparrow}$ & $\text{dev}_{dup}$ & $\text{dev}_{\neg dup}$ & $\text{PAWS}_{dup}$ & $\Delta^{\uparrow}$ & $\text{PAWS}_{\neg dup}$ & $\Delta^{\uparrow}$ \\
          \midrule
          ERM & \ding{51} & 0.842 & 0.834 &   -    & 0.615 &   -    & \textbf{0.883} & 0.915 & 0.852 &   -    & 0.232 & -  \\
          SR$\dagger$  & \ding{51} & 0.835 & 0.812 & -0.022 & 0.692 & +0.077 & 0.855 & 0.919 & 0.892 & +0.040 & 0.506 & +0.274  \\
          POE$\dagger$& \ding{51} & 0.829 & 0.810 & -0.024 & 0.679 & +0.064 & 0.808 & \textbf{0.935} &  0.710 & -0.142 & 0.499 & +0.267  \\
          CR$\dagger$ & \ding{51} & \textbf{0.845} & 0.827 & -0.007 & 0.691 & +0.076 & 0.855 & 0.915 & \textbf{0.910} & +0.058 & 0.198 & -0.034  \\
          \rowcolor[gray]{0.92}\textbf{MiMu} & \ding{55} & 0.832 & \textbf{0.838} &  +0.004
          & \textbf{0.707}  &  +0.092  & 0.882 & 0.930 & 0.698 & -0.154 & \textbf{0.546} & +0.314 \\
        \bottomrule
      \end{tabular}}
      \label{ex-nlp}
  \end{table*}

\subsubsection{Benchmark Methods}
\label{exp-methods}

Across all datasets, we first evaluate standard training that employs the ERM approach with pre-trained BERT \cite{bert} and ViT-Large \cite{vit} as network architectures to establish our baselines. 
Subsequently, we follow and align with current studies \cite{naacl21,acl23,liwhac-cvpr23} in NLP and CV fields, and conduct a comprehensive evaluation of shortcut mitigation methods. \textbf{In NLP}, we align with most studies \cite{naacl21,acl23} by choosing common baselines to alleviate shortcut learning, including Sample Re-weighting (SR), Product Of Experts (POE), and Confidence Regularization (CR). All the baselines use hand-crafted features that indicate how words are shared between the sentences pairs following \cite{ensemble}. However, MiMu is proposed to address the unknown and multiple shortcuts.

\begin{itemize}
    \item \textbf{Sample Re-Weighting} adjusts the weights of samples in the training set to ensure that the model does not overly focus on the biased examples. In the first step, a bias-only model $f_b$ that trained by the hand-crafted features is trained to measure how well the sample prediction given only the biased features. In the second step, the probability $p_b$ obtained by biased model is used to indicate shortcut degree and adjust the loss function of main model $f_d$, and $p_d$ is the probability outputs of $f_d$, the loss function can be defined as:
        \begin{equation}
            \mathcal{L} = - (1-{p_b}^c) \mathcal{L}_o(y, \hat{y}),
        \end{equation}
    where ${p_b}^c$ denotes ${p_b}$ on the correct label, model should more focus on samples of ${p_b}^c \rightarrow 0$, $\mathcal{L}_o$ can be a random original loss function.

    \item  \textbf{Product of Experts} integrates bias-only model $f_b$ to train main model $f_d$ by combining their softmax outputs $p_d$, $p_b$:
        \begin{equation}
            \mathcal{L} = - y \cdot \log softmax(\log{p_d}, \log{p_b}).
        \end{equation}

    \item  \textbf{Confidence Regularization} involves scaling the teacher model output $p_t$ using $p_b$ by biased model, which denotes as $S({p_b}^c, p_t)$. The main model $f_d$ is trained, and loss function could be defined as:
        \begin{equation}
            \mathcal{L} = - S({p_b}^c, p_t) \cdot \log{p_d}.
        \end{equation}

\end{itemize}

\textbf{In CV}, we follow the study \cite{liwhac-cvpr23,zhao2024comi} and adopt general data \textbf{augmentation and regularization} methods without prior knowledge of the shortcuts. These methods, commonly used to improve the accuracy of IC, include: Mixup \cite{cv-mixup}, Cutout \cite{cv-cutout}, CutMix \cite{cv-cutmix}, AugMix \cite{cv-augmix}, SD \cite{sd-nips21}. We also employ methods that infer the \textbf{pseudo shortcut labels} when ground-truth labels are unavailable, requiring two-stage training. These methods include LfF \cite{pseudo-LfF}, JTT \cite{pseudo-JTT}, EIIL \cite{pseudo-EIIL}, DebiAN \cite{pseudo-DebiAN}.

  \begin{table*}
    \centering
    \caption{The experimental results on the IN-9-W in CV. Baselines are categorized into three groups. }
    \resizebox{\textwidth}{!}{
    \begin{tabular}{c|c|c|ccccccc}
    \toprule
    \multirow{2}{*}{\textbf{Category}} & \multirow{2}{*}{\textbf{Methods}} & \multirow{2}{*}{\textbf{Multiple}} & \multicolumn{7}{c}{\textbf{IN-9-W}} \\
        ~ & ~ & ~ & dev & Rand-B & $\Delta^{\uparrow}$ & Rand-W & $\Delta^{\uparrow}$ & Rand-B+W & $\Delta^{\uparrow}$\\
      \midrule    
      \textbf{ERM} &  ViT-L \cite{vit} & - & 0.9796 & 0.7708 & -   & 0.7357 & -  & 0.7419 & -    \\
      \midrule
      \multirow{5}{*}{\textbf{\makecell{Augmentation \\ \& Regularization}}} & Mixup \cite{cv-mixup} & \ding{55}    & 0.9789 & 0.7684 &  -0.0024 &  0.7345 & -0.0012 & 0.7428 & +0.0009   \\
      ~ & Cutout \cite{cv-cutout} & \ding{55}    & 0.9804 & 0.7689 & -0.0019 & 0.7343 & -0.0014 & 0.7407 & -0.0012 \\
      ~ & CutMix \cite{cv-cutmix} & \ding{55}    & 0.9753 & 0.7684 & -0.0024 & 0.7307 & -0.0050 & 0.7371 & -0.0048  \\
      ~ & AugMix \cite{cv-augmix} & \ding{55}    & 0.9794 & 0.7670 & -0.0038  & 0.7355  & -0.0002  & 0.7364 & -0.0055 \\
      ~ & SD  \cite{sd-nips21}  & \ding{55}     & 0.9796 & 0.7708 & - & 0.7357 & - & 0.7416 & -    \\
      \midrule
      \multirow{4}{*}{\textbf{Pseudo Shortcut}} & LfF \cite{pseudo-LfF}  & \ding{55}   & 0.9792 & 0.7708 & -  & 0.7357 & - & 0.7409 & -0.0010 \\
      ~ & JTT \cite{pseudo-JTT} & \ding{55}      & 0.9806 & 0.7732 & +0.0024 & 0.7417 & +0.0060 & 0.7448 &  +0.0029  \\
      ~ & EIIL \cite{pseudo-EIIL} & \ding{55}     & 0.9811 & 0.7679 & -0.0029 & 0.7300 & -0.0057 & 0.7426 & +0.0007      \\
      ~ & DebiAN \cite{pseudo-DebiAN} & \ding{51}   & 0.9804 & 0.7710 & +0.0002 & 0.7384 & +0.0027 & 0.7450 & +0.0031 \\
      \midrule
      \rowcolor[gray]{0.92}\multirow{3}{*}{\textbf{Ours}} & \textbf{w/o s-c} & \ding{51} & 0.9780 & 0.7968 & +0.0260 & 0.7986 & +0.0629 & 0.7806  & +0.0387 \\
      \rowcolor[gray]{0.92}~ & \textbf{w/o s-i} & \ding{51} & 0.9751 & 0.7952 & +0.0244 & \textbf{0.8074} & +0.0717 & 0.7794 & +0.0375 \\
      \rowcolor[gray]{0.92}~ & \textbf{MiMu} & \ding{51} & 0.9782 & \textbf{0.8021} & +0.0313 & 0.8036 & +0.0679 & \textbf{0.7868}  & +0.0449\\
    \bottomrule
    \end{tabular}}
    \label{ex-cv-i9}
  \end{table*}

\subsubsection{Implementation Details}
Across all baselines, we employ pre-trained BERT \cite{bert} and VIT-Large \cite{vit} pre-trained on ImageNet-21k as our foundational architectures, i.e. transformer-based ERM models.
To evaluate the performance of these models, we use accuracy and top-1 accuracy metrics, the experiments conducted on the baselines are followed by the settings of \cite{acl23} and \cite{liwhac-cvpr23} in NLP and CV respectively. All experiments are implemented with Pytorch\footnote{https://pytorch.org/} on a Tesla A100 GPU.

\textbf{In NLP}, we employ BERT-base-uncased \footnote{https://huggingface.co/transformers} as ERM model. 
For optimization, we choose the AdamW optimizer \cite{adamw} with a learning rate range in [2e-5, 5e-5], and set the linear warmup ratio to 10\% of the total training episodes, use a batch size of 64, and train for 20 epochs. 
The max sequence length $l$ is set to 128, hidden size is 768.
\textbf{In CV}, we employ ViT-Large-Patch16-224-in21k\footnote{https://huggingface.co/google/vit-large-patch16-224-in21k} as ERM model. 
For image reprocessing, we scale and center-crop images to 224 $\times$ 224 with horizontal flipping and normalize to [-1, 1], sequence length $l$ is 197, hidden size is 1024. 
During training, we use stochastic gradient descent (SGD) optimizer \cite{sgd} with a learning rate range in [1e-3, 1e-4] and 1e-4 weight decay (i.e. $\ell_{2}$ penalty), and use a batch size of 64. All baselines are trained for 90 epochs.
In all settings, $N$ is range in [10\%, 20\%] of sequence length, and $M$ is the 50\% of top differences.
The relevant code could be obtained here\footnote{https://github.com/LiliizZ/MiMu}.

\subsection{Experimental Analysis}

\subsubsection{Main Performance Analysis}
\textbf{For NLP tasks}, we perform comparative experiments with common baselines, as presented in Table \ref{ex-nlp}. \underline{\textbf{On MNLI}}, compared to the ERM model, the comparative baselines typically achieve generalization enhancement at the expense of decreased accuracy of IID. For instance, the Product-Of-Expert (POE) improves the accuracy on Hans subset by 6.4\% but leads to a 2.4\% decline in the accuracy on mismatch subset.
However, MiMu enhances OOD generalization without compromising IID performance. Specifically, MiMu not only boosts its performance on the Hans dataset by 9.2\%, but also achieved a 0.4\% increase on mismatch.
This indicates that the self-calibration and self-improvement strategies can effectively mitigate multiple shortcut learning behavior.
\underline{\textbf{On QQP}}, baselines exhibit comparable behaviors on the IID dataset but diverge on the OOD-PAWS dataset, primarily attributed to the prevailing lexical overlap bias. Consequently, models frequently predict ``duplicated'' for most samples.
Among baselines, only Sample Re-weighting (SR) consistently performs well, maintaining a balance across both IID and OOD datasets. Although Confidence Regularization (CR) could perform well overall on MNLI, it exhibits low accuracy on ``non-duplicated'' subset of PAWS.
Unlike previous methods that rely on hand-crafted features, our method exhibits improvement and maintains a comparable level on both subsets of PAWS without known shortcuts. 
Overall, MiMu achieves a balanced performance and effectively mitigates strong spurious correlations between lexical overlap and ``duplicated''.

  \begin{table*}
    \centering
    \caption{The experimental results on the IN-200-W in CV. Baselines are categorized into three groups.}
    \resizebox{\textwidth}{!}{
    \begin{tabular}{c|c|c|ccccccc}
    \toprule
    \multirow{2}{*}{\textbf{Category}} & \multirow{2}{*}{\textbf{Methods}} & \multirow{2}{*}{\textbf{Multiple}} & \multicolumn{7}{c}{\textbf{IN-200-W}} \\
    ~ & ~ & ~ & dev & IN-B & $\Delta^{\uparrow}$ & IN-R & $\Delta^{\uparrow}$ & IN-R+W & $\Delta^{\uparrow}$ \\
        \midrule
        \textbf{ERM} &  ViT-L \cite{vit} & -  & 0.9122 & 0.6213 & - & 0.3876 & - & 0.3817 & -  \\
        \midrule
        \multirow{5}{*}{\textbf{\makecell{Augmentation \\ \& Regularization}}} & Mixup \cite{cv-mixup} & \ding{55} & 0.9102 & 0.6180 & -0.0033 & 0.3879 & +0.0003 & 0.3815 & -0.0002 \\
        ~ & Cutout \cite{cv-cutout} & \ding{55} & 0.9128 & 0.6197 & -0.0016 & 0.3874 & -0.0002 & 0.3816 & -0.0001 \\
        ~ & CutMix \cite{cv-cutmix} & \ding{55} & 0.8943 & 0.6063 & -0.0150 & 0.3773 & -0.0103 & 0.3719 & -0.0098 \\
        ~ & AugMix \cite{cv-augmix} & \ding{55} & 0.9130 & 0.6212 & -0.0001 & 0.3880 & +0.0004 & 0.3823 & +0.0006 \\
        ~ & SD \cite{sd-nips21}   & \ding{55}  & 0.9122 & 0.6213 & +0.0001 & 0.3876 & - & 0.3816 & -0.0001 \\
        \midrule
        \multirow{4}{*}{\textbf{Pseudo Shortcut}} & LfF \cite{pseudo-LfF} & \ding{55} & 0.8978 & 0.6086 & -0.0127 & 0.3800 & -0.0076 & 0.3733 & -0.0084 \\
        ~ & JTT \cite{pseudo-JTT} & \ding{55} & 0.9194 & 0.6259 & +0.0046 & 0.3944 & +0.0068 & 0.3881 & -0.0064 \\
        ~ & EIIL \cite{pseudo-EIIL} & \ding{55} & 0.8895 & 0.5730 & -0.0483 & 0.3250 & -0.0626 & 0.3196 & -0.0621 \\
        ~ & DebiAN \cite{pseudo-DebiAN} & \ding{51} & 0.9143 & 0.6211 & -0.0001 & 0.3867 & -0.0009 & 0.3813 & -0.0004 \\
        \midrule
        \rowcolor[gray]{0.92}\multirow{3}{*}{\textbf{Ours}} & \textbf{w/o s-c} & \ding{51} & 0.9146 & 0.6322 & +0.0109 & 0.4190 & +0.0314 & 0.4187 & - \\
        \rowcolor[gray]{0.92}~ & \textbf{w/o s-i} & \ding{51} & 0.9143 & 0.6382 & +0.0169 & 0.4149 & +0.0273 & 0.4149 & +0.0332 \\
        \rowcolor[gray]{0.92}~ & \textbf{MiMu} & \ding{51} & \textbf{0.9202} & \textbf{0.6400} & +0.0169 & \textbf{0.4254} & +0.0378 & \textbf{0.4214} & +0.0397 \\
    \bottomrule
    \end{tabular}}
    \label{ex-cv-i200}
  \end{table*}

\begin{table}[t]
  \caption{The experimental results of ablation study in NLP.}
  \begin{adjustbox}{width=\columnwidth}
  \centering
      \begin{tabular}{c|ccc|cccc}
      \hline
      \hline   
      \multirow{2}{*}{\textbf{Methods}} & \multicolumn{3}{c|}{\cellcolor[gray]{0.92}\textbf{MNLI}} & \multicolumn{4}{c}{\cellcolor[gray]{0.92}\textbf{QQP}} \\
        & \cellcolor[gray]{0.92}d-m & \cellcolor[gray]{0.92}d-mm & \cellcolor[gray]{0.92}Hans & \cellcolor[gray]{0.92}$\text{d}_{dup}$ & \cellcolor[gray]{0.92}$\text{d}_{\neg dup}$ & \cellcolor[gray]{0.92}$\text{P}_{dup}$ & \cellcolor[gray]{0.92}$\text{P}_{\neg dup}$ \\
        \midrule
        \textbf{MiMu} & 0.832 & 0.838 & \textbf{0.707} & 0.882 & 0.930 & 0.698 & \textbf{0.546}  \\
        w/o s-c  &  0.841 &  0.841 & 0.688 & 0.860 & 0.934 & 0.593 &  0.439 \\
        w/o s-i  & 0.839 & 0.840 & 0.655 & 0.900 & 0.917 & 0.821 & 0.210 \\
      \bottomrule
    \end{tabular}
    \end{adjustbox}
    \label{ablation_nlp}
    \vspace{-5pt}
\end{table}

\textbf{For CV tasks}, we perform comparative experiments with diverse baselines as shown in Tables \ref{ex-cv-i9} and \ref{ex-cv-i200}.
Our method, MiMu, outperforms all baselines, enhancing performance on OOD samples while maintaining excellent performance on IID samples. 
Due to the powerful pre-trained knowledge of ERM, 
all models perform similarly in IID sets, but with different degrees of decline in OOD sets.
Compared to VIT-L, simple standard augmentation and regularization methods may further rely on shortcuts when encounter multiple shortcuts, leading to decreased performance on OOD sets.
Methods of inferring pseudo shortcut labels are all based on the different assumptions that ERM learns different spurious correlations during training and exhibit poorer performance.
Among them, LfF infers shortcut labels by assuming that the shortcut is learned earlier. JTT and EIIL employ an under-trained ERM, trained for some epochs, as the reference model to infer pseudo shortcut labels. DebiAN trains the reference and mitigation models simultaneously. 
Nevertheless, model can learn different shortcuts at different stages, the assumption is too weak to contain all shortcuts and methods are inadequate to deal with varied shortcuts. 
Compared to them, MiMu performs well on OOD which contains single shortcut (Rand-B, Rand-W, IN-B, IN-R), which means combined strategies can mitigate the effects of multiple shortcuts in the training data. 
On OOD datasets containing two types of shortcuts compared to single shortcut, MiMu performs relatively well and significantly improves the performance of the model compared to baselines.

\begin{figure*}[t]
  \centering
  \includegraphics[width=\textwidth]{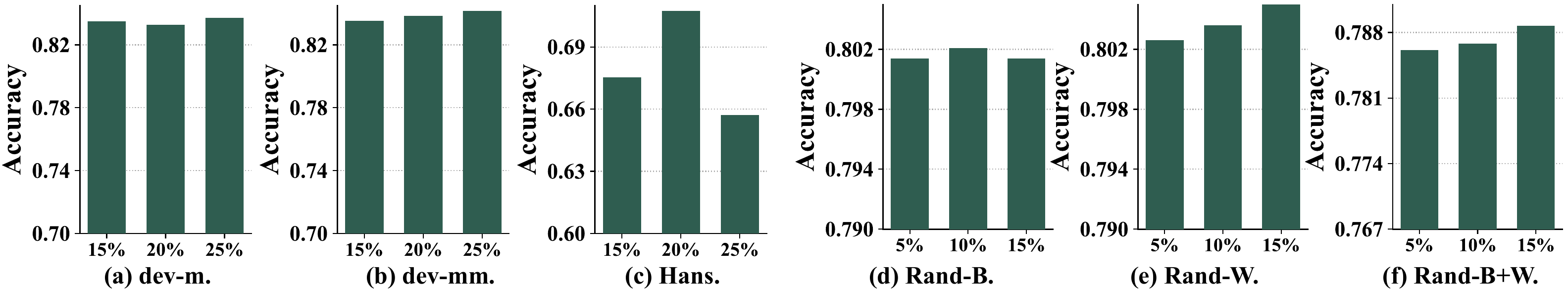}
  \caption{The experimental results on obscured position $N$ in MNLI (a,b,c) and IN-9-W (d,e,f).}
  \label{fig:k}
  \end{figure*}

  \begin{figure*}[t]
    \centering
    \includegraphics[width=\textwidth]{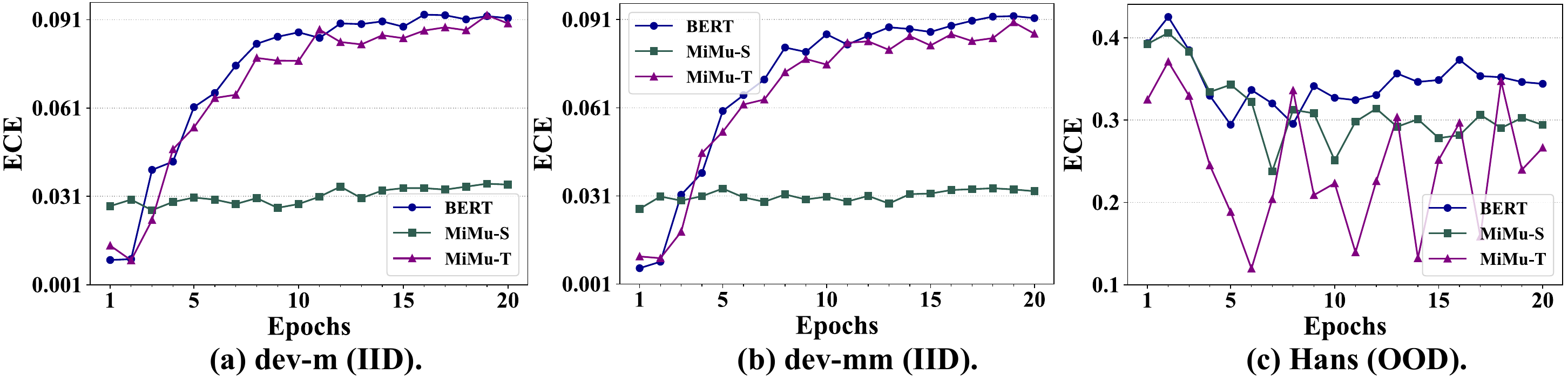} 
    \caption{The calibration experiments \cite{calibration} of ERM and MiMu on the IID and OOD samples of MNLI datasets. MiMu-S and MiMu-T represent $\mathcal{F}_s$ and $\mathcal{F}_t$. A lower ECE score indicates better calibration.}
    \label{ece}
  \end{figure*}

\subsubsection{Ablation Study} 
The experimental results of ablation studies in NLP are shown in Table \ref{ablation_nlp}, and the results in CV are shown in Tables \ref{ex-cv-i9} and \ref{ex-cv-i200}.
To evaluate the contributions of different components in MiMu, the self-calibration and self-improvement strategies are removed and denoted as ``w/o s-c'' and ``w/o s-i''. It can be seen that different modules play different roles in different tasks. 
\textbf{For NLP tasks}, in MNLI, removing self-calibration and self-improvement strategies leads to mixed outcomes. Specifically, the model's performance on OOD samples deteriorates, whereas its effectiveness on IID datasets shows a marginal increase. However, this increase is relatively modest when compared to the decrease in OOD performance. This phenomenon could be attributed to the roles of self-calibration and post-hoc calibration within the self-improvement strategy, which synergistically enhance the model's ability to handle OOD data significantly. Similarly, on QQP, the implementation of two strategies results in the model's performance stabilizing, effectively mitigating the impact of lexical overlap shortcuts.

\textbf{For CV tasks}, in IN-9-W, the removal of self-calibration and self-improvement strategies has led to a decline in the model's performance on both IID and OOD datasets. Notably, except for IN-9-W, the removal of the self-improvement strategy leads to a significant decrease in overall performance, indicating that this strategy effectively reduces multiple shortcuts, thereby enhancing the model's generalization capabilities. In Rand-W, the results of ``w/o s-i'' indicate that the $\mathcal{F}_s$ relies more on the strong shortcut-background than on the watermark. The introduction of self-improvement strategy makes it less dependent on strong shortcuts and finds a balance between strong and weak shortcuts.
In IN-200-W, in addition to textures and watermarks, we evaluate the impact of the background. It is clear that the model captures the spurious correlations between background and labels during training, and MiMu can find a balance between these shortcuts.

\subsubsection{Experiments on $N$}
In self-improvement strategy, to encourage a focus on broader locations rather than on a fixed region for learning shortcuts, we introduce a random masking strategy to selectively obscure $N$ positions. 
\textbf{Insufficient masked positions cause the model to focus on strong shortcuts, i.e. multiple shortcuts are not completely eliminated. Conversely, an excessive number of masked positions result in the model's lack of attention to the real underlying features, i.e. not learning well.} The experimental results regarding $N$ in MNLI and IN-9-W are shown in Fig. \ref{fig:k}. 
It can be observed that an inappropriate $N$ will lead to poor performance of the model in OOD settings. The best suitable $N$ is 20\% for tokens in NLP and 10\% for patches in CV.

\begin{figure*}[t]
  \centering
  \includegraphics[width=\textwidth]{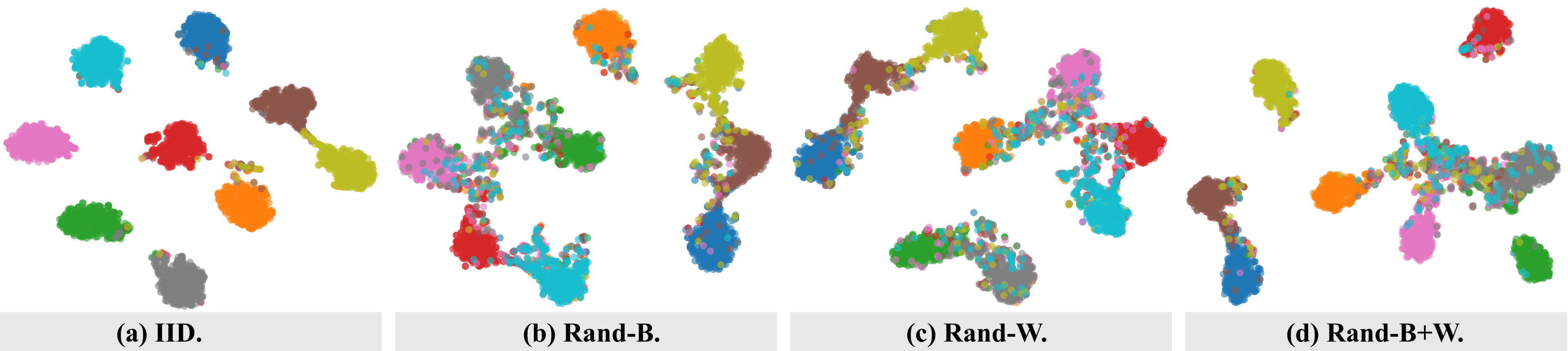}
  \caption{The embedding representations in IN-9-W with t-
  SNE visualization \cite{tsne}.}
  \label{tsne-i9}
\end{figure*}

\subsubsection{Calibration Analysis}
To better compare the calibration capabilities of ERM, source model $\mathcal{F}_s$, and target model $\mathcal{F}_t$ in MiMu, we employ Expected Calibration Error (ECE) \cite{calibration} scores by comparing the predicted probabilities with the actual accuracy of the predictions for evaluation.
\textbf{A lower ECE score indicates better calibration}, implying that the model's predicted probabilities align more closely with the actual accuracy of predictions. 
ECE can be computed as:
\begin{equation}
  \text{ECE} = \sum_{i=1}^{B} \frac{N_i}{N} \left| \text{accuracy}_i - \text{confidence}_i \right|,
\end{equation} 
where $B$ represents the number of probability bins, $N_i$ denotes the number of samples in the $i$-th probability bin, $N$ is the total number of samples, ${accuracy}_i$ stands for the observed accuracy for samples in the $i$-th probability bin, and ${confidence}_i$ indicates the average predicted probability for samples in $i$-th probability bin.

As shown in Fig. \ref{ece-i9} and \ref{ece}, certain findings could be obtained. First, the overall calibration of MiMu is better than that of ERM on IID datasets, among which $\mathcal{F}_s$ shows excellent calibration ability in each epoch, reflecting the superiority of self-calibration strategy.
Then, on the OOD dataset, each model exhibits fluctuation in ECE scores, in which the ECE scores of $\mathcal{F}_t$ are lower on the whole. Compared with $\mathcal{F}_s$, $\mathcal{F}_t$ shows stronger calibration ability, highlighting the superiority of self-improvement strategy.

\subsubsection{Embedding representation Analysis}
As shown in Fig. \ref{tsne-i9}, we employ t-SNE \cite{tsne} visualization to represent embeddings on the IN-9-W dataset with four validation subsets: IID, Rand-B, Rand-W, Rand-B+W. The representations naturally segregate into 9 clusters in the embedding space, reflecting the 9 classifications in the ImageNet-9 task. Consistent with the experimental results shown in Table \ref{ex-cv-i9}, on Rand-W, the embedding representations of 9 categories are best separated, while on Rand-B+W, the representations of the blue category and the gray category are somewhat indistinct.

\begin{figure*}[t]
  \centering
  \includegraphics[width=\textwidth]{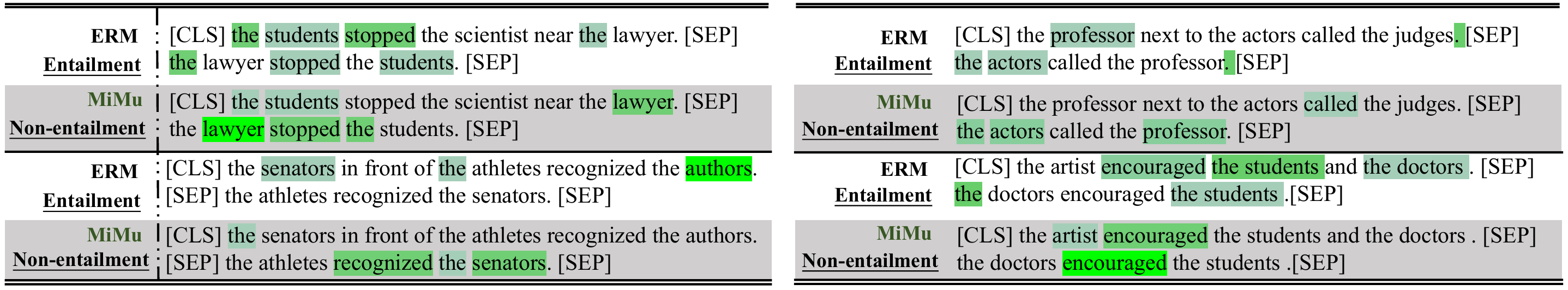}
  \caption{The saliency visualization in Hans predicted by ERM and MiMu, the \underline{underline} represents the prediction results of models (ERM, MiMu), and the right part of each single figure consists of integrated gradient vectors \cite{integrated-gradient} of tokens, darker colors indicate higher attention.}
  \label{sali_nlp}
\end{figure*}

\begin{figure*}[t]
  \centering
  \includegraphics[width=\textwidth]{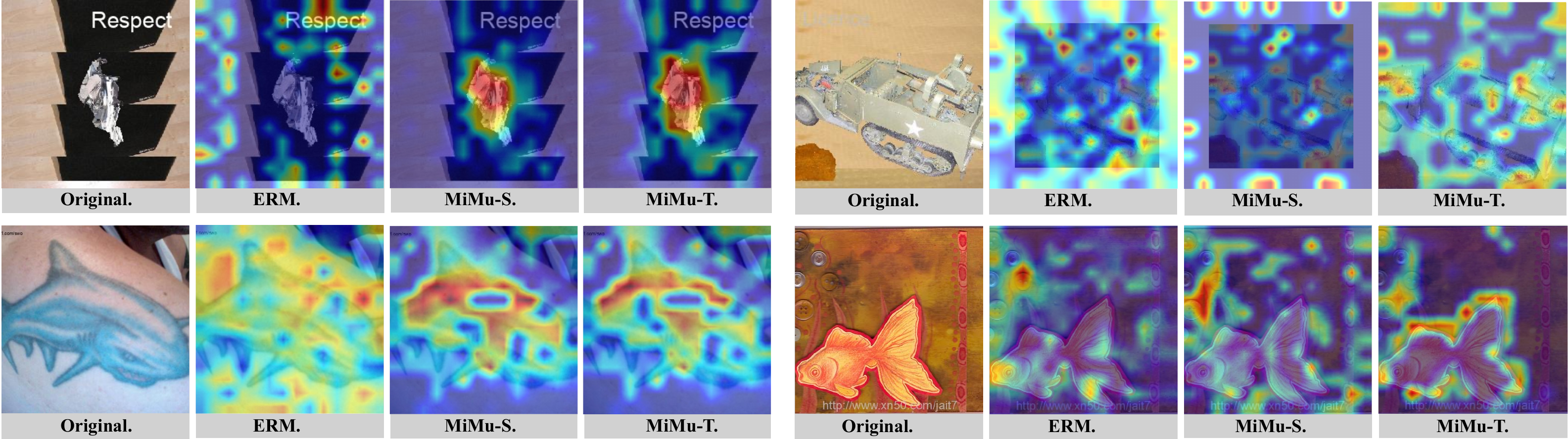} 
  \caption{The saliency maps \cite{saliency} in Rand-B+W (top line) and IN-R+W (bottom line). From left to right in each case are from the original image, the pre-trained ViT-Large. MiMu-S and MiMu-T represent $\mathcal{F}_s$ and $\mathcal{F}_t$, warm colors indicate a high level of attention for models.}
  \label{sali_cv-i9-i200}
\end{figure*}

\subsubsection{Saliency Analysis}
\textbf{For NLP tasks}, comparative cases between ERM and MiMu via integrated gradient explanation \cite{integrated-gradient} are presented in Fig. \ref{sali_nlp}.
The cases are from Hans subset of MNLI, due to the word overlap bias in dataset, ERM often determines the relationship between two sentences as ``entailment'' but MiMu makes right prediction. 
In addition to word overlap, we find that ERM is more likely to focus on nonsense words such as symbols (e.g. ``.''), articles (e.g. ``the''), and focus on only premise ignoring hypothesis in the second case.
In contrast to ERM, MiMu focuses more on the relationship between premise and hypothesis. 
In comparison, MiMu pays less attention to meaningless articles and focuses more on meanings of sentences themselves.

\textbf{For CV task}, we adopt Grad-CAM \cite{saliency} to present saliency maps of some images to evaluate the performance on alleviating multiple shortcuts. As illustrated in Fig. \ref{sali_cv-i9-i200}, the pre-trained ViT-Large (i.e. ERM) fails in capturing real objects and makes false predictions. MiMu performs adeptly by accurately detecting real objects. 
It can be seen that calibration loss in $F_s$ (MiMu-S) initially enriches feature distribution, resulting in focusing on more locations. The self-improvement strategy in $F_t$ (MiMu-T) further forces the model to focus on true underlying features by random mask strategy and adaptive attention alignment module.

\section{Discussions}

\label{append:limit}
MiMu has achieved excellent experimental results. However, during the experiments, we also find some limitations and work needed to be done in the future.

\textbf{Unknown Shortcuts:} The information contained in images or texts is very extensive, and how models exploit vulnerabilities and take shortcuts still requires further attention. For example, text may contain more than lexical overlap, subsequence overlap;images may contain more than watermarks, backgrounds, textures, and other shortcuts. Especially in Large Language Models (LLMs) \cite{yuan2024llms,zhao2025evaluating}, there are diverse shortcuts that LLMs can capture, so LLMs hallucinate when faced with common problems, which still needs further efforts.
\textbf{In this paper, we aim at mitigating the influence of shortcut learning, not eliminating.} 
In theory, it is impossible to completely eliminate the effect of shortcuts, and even when trained on images of individual objects, the model cannot perform optimally on OOD sets as shown in Fig. \ref{fig:ei}, because the model may capture the correlations between the color, certain positions, and labels as shortcuts to predict.

\textbf{Corse-grained Shortcut:} The current research of shortcut learning behavior is coarse-grained. In fact, more fine-grained research is needed, it may not be limited to the spurious correlation between carpet background and pet dogs. This background requires more detailed research, such as the size, the complexity, the color, etc. Similarly, the style, semantic information and location of the watermark will affect the degree of shortcut learning behavior of the model. These fine-grained studies could help us better understand how deep learning systems learn shortcuts and improve the robustness generalization abilities of learning systems.

\textbf{The interaction of multiple shortcuts:} As shown in bottom row of Fig. \ref{fig:ei} and Assumption \ref{ass:2}, when two types of shortcuts are combined, the model's performance on OOD datasets always falls between the model's performance on each individual shortcut. 
In addition to the possible presence of other shortcuts in the data, we hypothesize several potential reasons as follows:
\begin{itemize}
    \item The combination of strong and weak shortcuts results in the model relying less on the strong shortcuts, making the model's performance in-between.
    \item The model also captures the correlation between the strong and the weak shortcuts, i.e., the model learns that the third shortcut. The specific correlations learned by the model need the wider research.
\end{itemize}
Besides, We find MiMu can help the model perform well on OOD datasets containing single shortcut, but its performance on multiple shortcuts is not as effective as on a single shortcut. We speculate that this may be because MiMu focuses on mitigating shortcut at the pixel or token level, while a token may contain information from multiple shortcuts. \textbf{The third type of shortcut captured by the model itself may be the interactions between shortcuts, which could be semantic relevance or combination relationships.} 
The effect of the interactions between the multiple shortcuts on the model are unknown and require further study.
We will study the impact of the interaction effect of multiple shortcuts on ERM models in the future work.

\section{Conclusions}

In this paper, we proposed a novel method (\textbf{MiMu}) with Transformer-based ERMs to \textbf{Mi}tigate \textbf{Mu}ltiple shortcut learning behavior, improving robustness generalization capabilities of learning systems. Specifically, we first introduced self-calibration strategy in source model to  prevent the model from being overconfident on shortcuts. Then, we designed self-improvement strategy to further address the problem that eliminating weak shortcuts while retaining strong ones during the process of eliminating shortcuts. 
Experimental results on four datasets in NLP and CV clearly demonstrated the effectiveness of our proposed method.

\begin{acknowledgement}
  This research was supported by grants from the National Key Research and Development Program of China (Grant No. 2024YFC3308200), the Key Technologies R \& D Program of Anhui Province (No. 202423k09020039), and the Fundamental Research Funds for the Central Universities.
\end{acknowledgement}


\section*{Appendix}
\label{appendix}
\subsection*{\textbf{Dataset}}
\label{append:data}

\begin{figure*}[!t]
    \centering
    \includegraphics[width=0.9\textwidth]{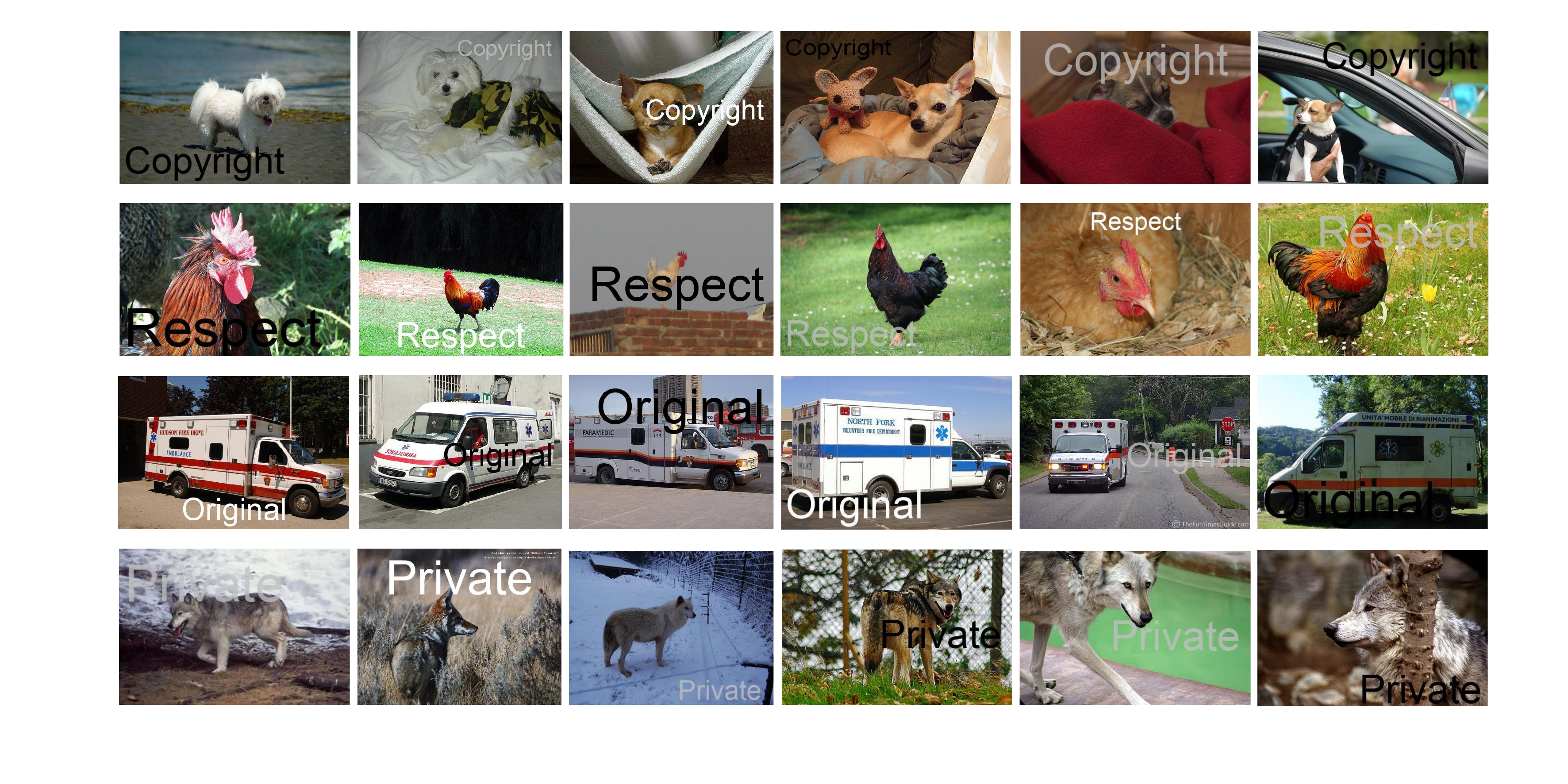}
    \caption{The varied styles of watermarks in IN-9-W, each type of watermark uses a fixed word indicating copyright protection. The size, position and color of watermarks are different. We focus on whether the model captures spurious relationships between the watermark content and class.}
    \label{appen:style_wm_9}
\end{figure*}

\begin{figure*}[!t]
    \centering
    \includegraphics[width=0.9\textwidth]{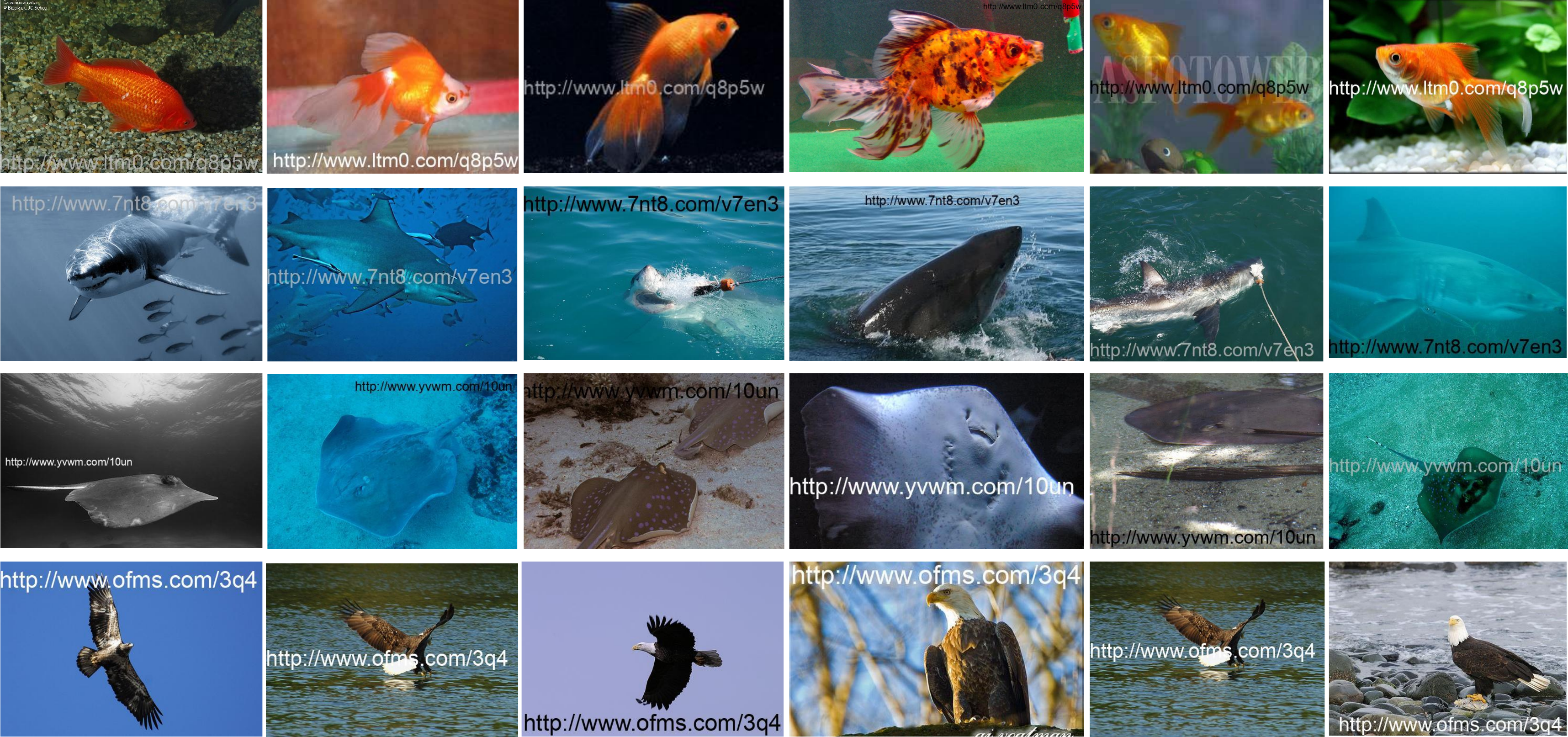} 
    \caption{The varied styles of watermarks on IN-200-W, each type of watermark uses a fixed web URL simulating real-world scenarios. In the OOD sets, we add random URLs of other classes.}
    \label{appen:style_wm_200}
\end{figure*}

For NLP task, we list multiple shortcuts model may rely on as presented in Table \ref{appen:vary-shortcut}.
For CV task, We adopt two watermarking methods that would appear in real scenes to fit the complexity of actual network data. 
\textbf{In IN-9-W}, we append different watermarks on different categories as shown in Fig. \ref{appen:style_wm_9}, these watermarks are related to copyright protection that appear in the actual situation and would not affect the image semantics, such as \textit{Ownership, Original, License, Allowed, Respect, Private, Watermark, Protect, Copyright} in Fig. \ref{appen:style_wm_9}.
\textbf{In IN-200-W}, since the network data is often marked with the URL watermark, we simulate web URL watermarking in real data on ImageNet-200 as presented in Fig. \ref{appen:style_wm_200}.
Since watermarks on each category are different, we generate 200 random web URLs, each comprising a base string, two randomly generated strings of 3 to 5 characters, and standard URL components, such as \textit{``http://www.7xd9c.com/0u1z''}.

\begin{table*}
    \centering
    \caption{The varied shortcuts of Hans dataset \cite{hans}. ``Non'' represents Non-entailment label.}
    \resizebox{0.8\textwidth}{!}{
    \begin{tabular}{l|lll}
        \toprule
        Shortcut & Premise & Hypothesis & Label \\
        \midrule
        \multirow{2}*{Lexical Overlap} & The doctors visited the lawyer. & The lawyer visited the doctors. & Non \\
        ~ & The judge by the actor stopped the banker. & The banker stopped the actor. & Non \\
        \midrule
        \multirow{2}*{Subsequence} & The senator near the lawyer danced. & The lawyer danced. & Non \\
        ~ & The judges heard the actors resigned. & The judges heard the actors. & Non \\
        \midrule
        \multirow{2}*{Constituent} & If the actor slept, the judge saw the artist. & The actor slept. & Non \\
        ~ & The lawyers resigned, or the artist slept. & The artist slept. & Non \\
        \bottomrule
    \end{tabular}}
    \label{appen:vary-shortcut}
    \vspace{10pt}
\end{table*}

\begin{figure}[t]
    \centering
    \includegraphics[width=0.48\textwidth]{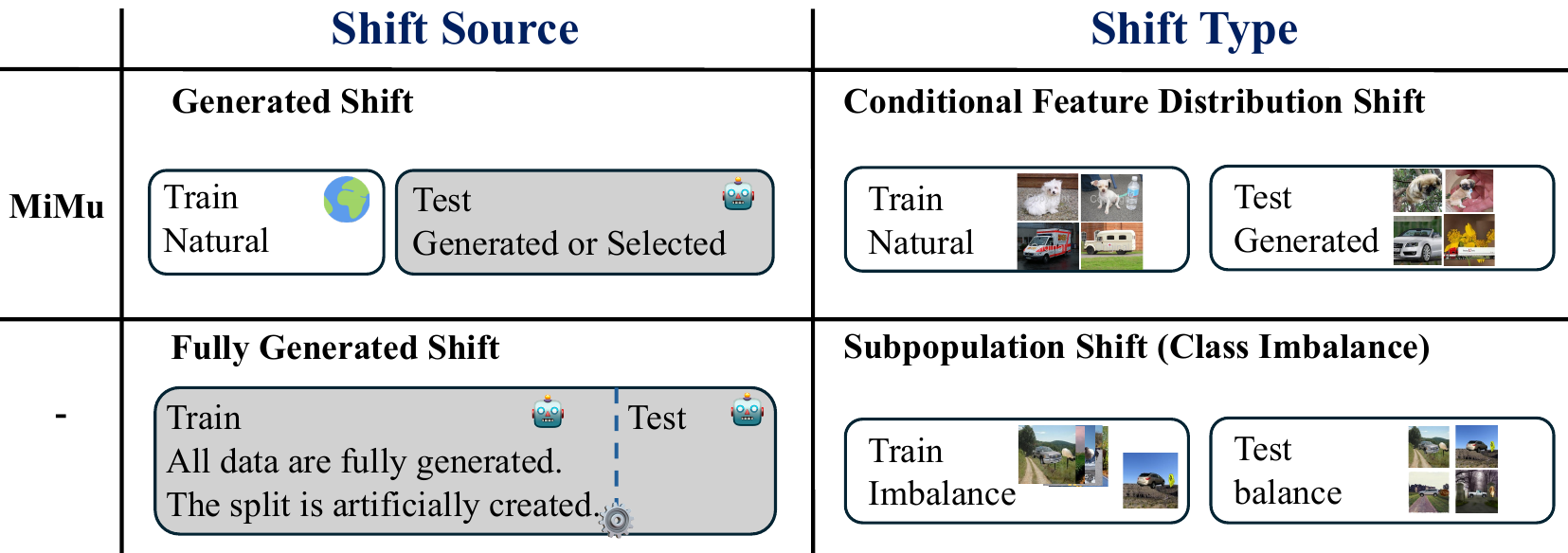} 
    \caption{The comparisons of shift source and shift type in MiMu and some research in CV.}
    \label{appen:sldebias}
\end{figure}

\subsection*{\textbf{Shifts in Shortcut Learning}}
\label{append:sl-debias}
As highlighted in \cite{hupkes2023taxonomy}, \textbf{the motivation of a study determines what type of generalization is desirable and shapes its experimental design.}
In this paper, we concentrate on shortcut learning, a specific generalization issue, arises from model inductive biases \cite{NMI20}. \textbf{Since the shortcuts captured by the model in the training are unknown, it can only be evaluated by post-hoc evaluation, we rigorously unify the shift of the datasets in NLP and CV.}
As mentioned in Section \ref{sec:rw}, to unify the settings, we focus more on generated shifts rather than fully generated shifts. As depicted in Fig. \ref{appen:sldebias}, we train models on natural data and make evaluations on generated OOD sets to assess which shortcuts the models rely on and to what extent the models rely on shortcuts. 
In some works \cite{asgari2022masktune,bg-eccv18,ensemble} and UrbanCars \cite{liwhac-cvpr23} constructed in CV are fully generated data, the train and test sets split is artificially created, which is against that the shortcuts captured by the model in the training are unknown. Meanwhile, it does not occur in NLP datasets.

Besides, as emphasized in Section \ref{sec:rw} and \ref{sec:pd}, models capture spurious correlations between features and labels in training, when the spurious correlations are broken in test, the model fails to perform well. Therefore, we focus on conditional feature distribution shift, which could be defined as $P_{tr}(X | Y) \neq P_{te}(X | Y)$ and $P_{tr}(Y) = P_{te}(Y)$.  We make some comparison with other shift types as follows:

\begin{itemize}
    \item Label distribution shift: $P(Y)$ might change but the class-conditional distribution $P(X|Y)$ do not: $P_{tr}(X | Y) = P_{te}(X | Y)$, $P_{tr}(Y) \neq P_{te}(Y)$.
    \item Semantic shift: In different environments or times $t$ and $t'$, for the same data x, $P_{t}(Y | X=x) \neq P_{t'}(Y | X=x)$ \\ $P_{t}(Y) \neq P_{t'}(Y)$, which indicates the semantics of data changes over time, place, or other external factors.
    \item Conditional feature distribution shift: $P_{tr}(X | Y) \neq P_{te}(X | Y)$ and $P_{tr}(Y) = P_{te}(Y)$, which indicates label distribution $P(Y)$ doesn't change, but the conditional distribution of the feature $X$ at the given $Y$ is changed.

\end{itemize}

\bibliographystyle{fcs}
\bibliography{sample-base}

\end{document}